\pdfoutput=1

\documentclass[11pt]{article}

\usepackage[]{EMNLP2022}

\usepackage{times}
\usepackage{latexsym}
\usepackage{amsmath,amsfonts,amssymb}
\usepackage[ruled,vlined]{algorithm2e}

\usepackage[T1]{fontenc}

\usepackage[utf8]{inputenc}

\usepackage{microtype}

\usepackage{inconsolata}

\definecolor{light_blue}{HTML}{DCE6F1}
\usepackage[most]{tcolorbox}
\tcbset{on line, 
        boxsep=1pt, left=0pt,right=0pt,top=0pt,bottom=0pt,
        colframe=white,colback=light_blue,  
        highlight math style={enhanced}
        }
        
\usepackage{caption}
\usepackage{graphicx}
\usepackage{booktabs}
\usepackage{xcolor}
\usepackage{multirow}
\usepackage{subcaption}
\usepackage[shortlabels]{enumitem}

\newcommand\numberthis{\addtocounter{equation}{1}\tag{\theequation}}

\newcommand{\vv}{$\mathbf{v}$}

\newcommand{\myhat}[1]{\mathbf{\tilde{\text{$#1$}}}}

\newcommand\wikilm{\textsc{WikiLM}}
\newcommand\gpt{\textsc{GPT2}}
\newcommand\wordf{\textsc{WordFilter}}
\newcommand\realtoxic{\textsc{RealToxicPrompts}}
\newcommand\wikitext{\textsc{WikiText-103}}

\title{Transformer Feed-Forward Layers Build Predictions by\\Promoting Concepts in the Vocabulary Space}

\author{Mor Geva\thanks{\hspace*{5px}Equal contribution.}$^{*,1}$ ~~~~~ Avi Caciularu$^{*,2,}$\thanks{\hspace*{5px}Work done during an internship at AI2.} ~~~~~ Kevin Ro Wang $^{3}$ ~~~~~
Yoav Goldberg$^{1,2}$ 
\vspace{0.2cm} \\
$^1$Allen Institute for AI ~~~ $^2$Bar-Ilan University ~~~ $^3$Independent Researcher \\
\small{\texttt{morp@allenai.org}},\small{\texttt{\{avi.c33,kevinrowang,yoav.goldberg\}@gmail.com}}}

\begin{document}
\maketitle

\begin{abstract}

Transformer-based language models (LMs) are at the core of modern NLP, but their internal prediction construction process is opaque and largely not understood. In this work, we make a substantial step towards unveiling this underlying prediction process,  by reverse-engineering the operation of the feed-forward network (FFN) layers, one of the building blocks of transformer models.
We view the token representation as a changing distribution over the vocabulary, and the output from each FFN layer as an additive update to that distribution. Then, we analyze the FFN updates in the vocabulary space, showing that each update can be decomposed to sub-updates corresponding to single FFN parameter vectors, each promoting concepts that are often human-interpretable. 
We then leverage these findings for controlling LM predictions, where we reduce the toxicity of \gpt{} by almost 50\%, and for improving computation efficiency with a simple early exit rule, saving 20\% of computation on average.\footnote{Our codebase is available at \url{https://github.com/aviclu/ffn-values}.}

\end{abstract}

\section{Introduction}
\label{sec:intro}
How do transformer-based language models (LMs) construct predictions? We study this question through the lens of the feed-forward network (FFN) layers, one of the core components in transformers \cite{vaswani2017attention}.
Recent work showed that these layers play an important role in LMs, acting as memories that encode factual and linguistic knowledge \cite{geva-etal-2021-transformer, da2021analyzing, meng2022locating}.
In this work, we investigate how outputs from the FFN layers are utilized internally to build predictions.

\begin{figure}[t]
    \centering
    \includegraphics[scale=0.74]{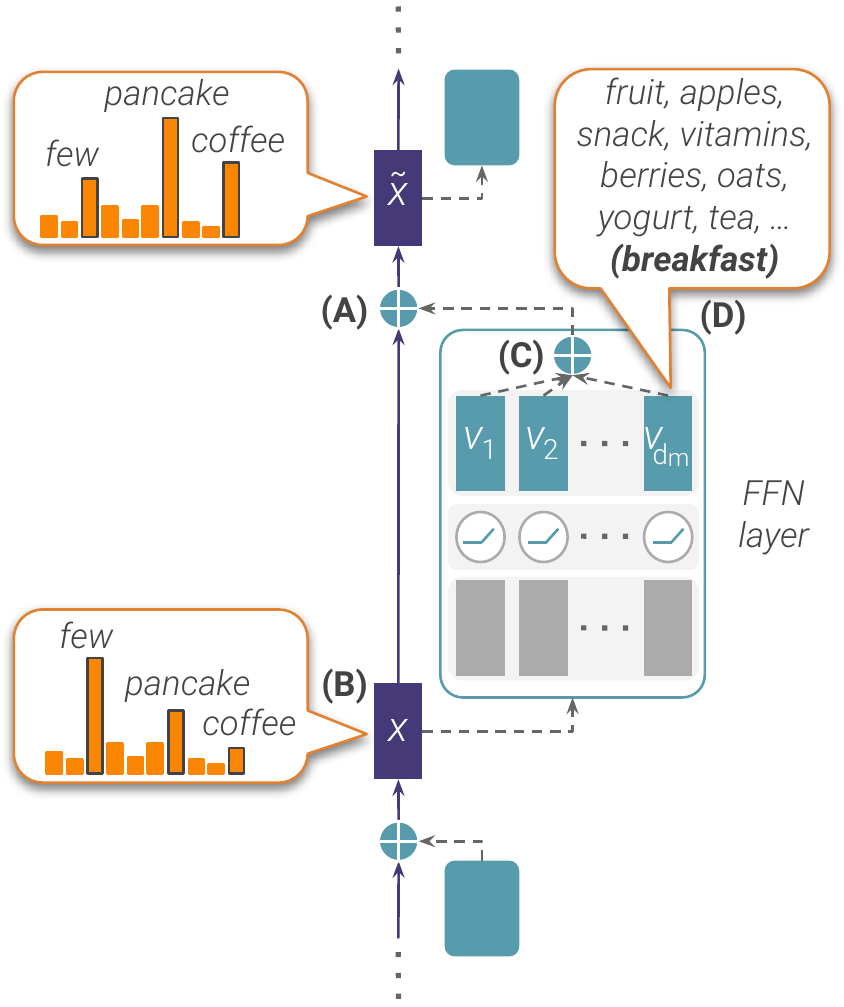}
    \caption{Illustration of our findings. Feed-forward layers apply additive updates (A) to the token representation $\mathbf{x}$, which can be interpreted as a distribution over the vocabulary (B). An update is a set of sub-updates induced by parameter vectors $\mathbf{v}_1,...,\mathbf{v}_{d_m}$ (C), each can be interpreted as a concept in the vocabulary space (D).}
    \label{figure:intro}
\end{figure}

We begin by making two observations with respect to the representation of a single token in the input, depicted in Fig.~\ref{figure:intro}. First, each FFN layer induces an additive update to the token representation (Fig.~\ref{figure:intro},\textbf{A}).
Second, the token representation across the layers can be translated \emph{at any stage} to a distribution over the output vocabulary \cite{geva-etal-2021-transformer} (Fig.~\ref{figure:intro},\textbf{B}).
We reason that the additive component in the update changes this distribution (\S\ref{sec:preliminaries}), namely, \textit{FFN layers compute updates that can be interpreted in terms of the output vocabulary}.

We then decompose the FFN update (\S\ref{sec:hypothesis}), interpreting it as a collection of sub-updates, each corresponding to a column in the second FFN matrix (Fig.~\ref{figure:intro},\textbf{C}) that scales the token probabilities in the output distribution.
Through a series of experiments, we find that (a) sub-update vectors across the entire network often encode a small-set of human-interpretable well-defined concepts, e.g. \textit{``breakfast''} or \textit{``pronouns''} (\S\ref{sec:values_concepts}, Fig.~\ref{figure:intro},\textbf{D}), and (b) FFN updates rely primarily on token promotion (rather than elimination), namely, tokens in the top of the output distribution are those pushed strong enough by sub-updates (\S\ref{sec:values_promotion_fixation}).
Overall, these findings allow fine-grained interpretation of the FFN operation, providing better understanding of the prediction construction process in LMs.

Beyond interpretation, our findings also have practical utility. In \S\ref{sec:toxic_language_suppression}, we show how we can intervene in the prediction process, in order to manipulate the output distribution in a direction of our choice. Specifically, we show that increasing the weight of only 10 sub-updates in \gpt{} reduces toxicity in its generations by almost 50\%. 
Also, in \S\ref{sec:early_exit}, we show that dominant sub-updates provide a useful signal for predicting an early exit point, saving 20\% of the computation on average.

In conclusion, we investigate the mechanism in which FFN layers update the inner representations of transformer-based LMs. We propose that the FFN output can be viewed as a collection of updates that promote concrete concepts in the vocabulary space, and that these concepts are often interpretable for humans. 
Our findings shed light on the prediction construction process in modern LMs, suggesting promising research directions for interpretability, control, and efficiency.

\section{Token Representations as Evolving Distributions Over the Vocabulary}
\label{sec:preliminaries}

Modern LMs \cite{baevski2018adaptive,radford2019language,NEURIPS2020_1457c0d6} are transformer models primarily trained to predict the next token probability for a given input.
Such LMs are composed of intertwined multi-head self-attention (MHSA) layers and FFN layers \cite{vaswani2017attention}, with residual connections \cite{he2016deep} between each pair of consecutive layers. The LM prediction is obtained by projecting the output vector from the final layer to an embedding matrix $E\in\mathbb{R}^{|\mathcal{V}|\times d}$, with a hidden dimension $d$, to get a distribution over a vocabulary $\mathcal{V}$ (after softmax).

Given a sequence ${\mathbf{w} = \langle w_1, ..., w_t \rangle}$ of input tokens, the model creates a contextualized representation ${\mathbf{x}_i \in\mathbb{R}^d}$ for each token ${w_i \in \mathbf{w}}$, that is being updated throughout the layers.
In this work, we analyze the updates applied by the FFN layers and how they construct the model prediction. 
Concretely, each FFN layer $\ell = 1, ..., L$ processes $\mathbf{x}_i^\ell$ and produces an output $\mathbf{o}_i^\ell$, which is then added to $\mathbf{x}_i^\ell$ to yield an updated representation $\myhat{\mathbf{x}}_i^\ell$:
\begin{align*}
    \mathbf{o}_i^\ell &= \texttt{FFN}^\ell(\mathbf{x}_i^\ell) \\
    \myhat{\mathbf{x}}_i^\ell &= \mathbf{x}_i^\ell+ \mathbf{o}_i^\ell
\end{align*}
The updated representation $\myhat{\mathbf{x}}_i^\ell$ then goes through a MHSA layer,\footnote{In some LMs, e.g. \gpt{}, a layer normalization (LN) \cite{ba2016layer} is applied to the representation $\myhat{\mathbf{x}}_i^\ell$. We omit it here and show it does not influence our interpretation in \S\ref{sec:hypothesis}.} yielding the input $\mathbf{x}_i^{\ell+1}$ for the next FFN layer.
The evolving representation in this process (i.e. $\mathbf{x}_i^\ell \rightarrow \myhat{\mathbf{x}}_i^\ell, \,\forall \ell$) can be viewed as an information stream that is being processed and updated by the layers \cite{elhage2021mathematical}. The output probability distribution is obtained from the final representation of the token, i.e.,
\begin{align}
\label{eq:proj}
    \mathbf{y} = \text{softmax}(E \myhat{\mathbf{x}}_i^L).
\end{align}

To analyze the FFN updates, we read from the representation \textit{at any layer} a distribution over the output vocabulary, by applying the same projection as in Eq.~\ref{eq:proj} \cite{geva-etal-2021-transformer}:
\begin{align*}
    \mathbf{p}_i^{\ell} &= \text{softmax}(E \mathbf{x}_i^{\ell}) \\
    \myhat{\mathbf{p}}_i^{\ell} &= \text{softmax}(E \myhat{\mathbf{x}}_i^{\ell}).
\end{align*}
Note that $\myhat{\mathbf{p}}_i^{L}=\mathbf{y}$.
Importantly, by linearity:
$$E \myhat{\mathbf{x}}_i^\ell = E \mathbf{x}_i^\ell + E \mathbf{o}_i^\ell,$$ implying that $\mathbf{o}_i^\ell$ can be interpreted as an additive update in the vocabulary space.
However, we find that the projection of the FFN output $E \mathbf{o}_i^\ell$ to the vocabulary is not interpretable (\S\ref{sec:values_concepts}). 
In this work, we take this a step further, and decompose the update $\mathbf{o}_i^\ell$ into a set of smaller sub-updates. By projecting the sub-updates to the vocabulary we find that they often express human-interpretable concepts.

In the rest of the paper, we focus on FFN updates to the representation of a single token in the sequence, and omit the token index for brevity, i.e. $\mathbf{x}^{\ell} := \mathbf{x}_i^{\ell}$ and $\mathbf{p}^{\ell} := \mathbf{p}_i^{\ell}$.

\section{The FFN Output as a Collection of Updates to the Output Distribution}
\label{sec:hypothesis}

We now decompose the FFN output, and interpret it as a set of sub-updates in the vocabulary space.

\paragraph{FFN Outputs as Linear Vector Combinations.} 
Each FFN at layer $\ell$ consists of two linear transformations with a point-wise activation function in between (bias terms are omitted):
$$
  \texttt{FFN}^\ell(\mathbf{x}^{\ell}) = f\left(W_K^\ell \mathbf{x}^{\ell} \right) W_V^\ell,
$$
where $W_K^\ell, W_V^\ell \in \mathbb{R}^{d_m \times d}$ are parameter matrices, and $f$ is a non-linearity function.
Previous work proposed this module can be cast as an emulated neural key-value memory~\cite{sukhbaatar2015end,sukhbaatar2019}, where rows in $W_K^\ell$ and columns in $W_V^\ell$ are viewed as keys and values, respectively \cite{geva-etal-2021-transformer}. For an input $\mathbf{x}^{\ell}$, the keys produce a vector of coefficients $\mathbf{m}^\ell := f\left(W_K^\ell \mathbf{x}^{\ell} \right) \in \mathbb{R}^{d_m}$, that weighs the corresponding values in $W_V^\ell$.
Denoting by $\mathbf{k}_i^{\ell}$ the $i$-th row of $W_K^{\ell}$ and by $\mathbf{v}_i^{\ell}$ the $i$-th column of $W_V^{\ell}$, we can then use the following formulation:
$$
  \texttt{FFN}^\ell(\mathbf{x}^{\ell}) = \sum_{i=1}^{d_m} f(\mathbf{x}^{\ell} \cdot \mathbf{k}_i^{\ell}) \mathbf{v}_i^{\ell} = \sum_{i=1}^{d_m} m_i^{\ell} \mathbf{v}_i^{\ell}.
$$
Therefore, \textit{a FFN update can be viewed as a collection of sub-updates, each corresponding to a weighted value vector in the FFN output.}

\paragraph{Terminology.} In the rest of the paper, we refer to the vectors $\mathbf{v}_i^{\ell}$ as \emph{value vectors}, and to their weighted form $m_i^\ell \mathbf{v}_i^{\ell}$ as \emph{sub-updates}. A transformer LM with $L=10, d_m=3000$ will have $30,000$ value vectors, and every token that passes through the transformer will weight these value vectors differently, resulting in $30,000$ sub-updates, where only a few of the sub-updates have high weights.

\paragraph{Interpreting Sub-Updates in the Vocabulary Space.}
Consider a sub-update $m_i^{\ell}\mathbf{v}_i^{\ell}$ for a given input, we can estimate its influence on the representation $\mathbf{x}^{\ell}$ (before the FFN update) by analyzing the change it induces on the output distribution. 
Concretely, we isolate the effect of $m_i^{\ell}\mathbf{v}_i^{\ell}$ on the probability $\mathbf{p}^{\ell}_w$ of $w \in \mathcal{V}$:\footnote{As in Eq.\ref{eq:proj}, LN is omitted. In App.~\ref{sec:value_layer_norm}, we verify empirically that our findings hold also when LN is applied.}
\begin{align*}
\label{eq:pred}
  p&\big(w \;|\; \mathbf{x}^{\ell} + m_i^{\ell}\mathbf{v}_i^{\ell}, E\big) \\
  &= \frac{\exp{\big(\mathbf{e}_w\cdot\mathbf{x}^{\ell}+\mathbf{e}_w\cdot m_i^{\ell}\mathbf{v}_i^{\ell}\big)}}{Z\big(E (\mathbf{x}^{\ell} + m_i^{\ell}\mathbf{v}_i^{\ell}))} \\
  &\propto \exp{\big(\mathbf{e}_w\cdot\mathbf{x}^{\ell}\big)} \cdot \exp{\big(\mathbf{e}_w \cdot m_i^{\ell}\mathbf{v}_i^{\ell}\big)}, \numberthis
\end{align*}
where $\mathbf{e}_w$ is the embedding of $w$, and $Z\big(\cdot\big)$ is the constant softmax normalization factor. 

This implies that each sub-update ${m_i^{\ell}\mathbf{v}_i^{\ell}}$ introduces a scaling factor to the probability of every token $w$ based on its dot product with $\mathbf{e}_w$.
Specifically, having ${\mathbf{e}_w \cdot m_i^{\ell}\mathbf{v}_i^{\ell} > 0}$ increases the probability of $w$, and having ${\mathbf{e}_w \cdot m_i^{\ell}\mathbf{v}_i^{\ell} < 0}$ decreases it.
This scaling factor can be split into two parts:
\begin{itemize}
[leftmargin=*,topsep=3pt,itemsep=3pt,parsep=0pt]
  
    \item The term ${\mathbf{e}_w\cdot\mathbf{v}_i^{\ell}}$ can be viewed as a \textit{static score} of $w$ that is independent of the input to the model. Thus, the projection ${\mathbf{r}_i^\ell = E \mathbf{v}_i^\ell \in \mathbb{R}^{|\mathcal{V}|}}$ induces a ranking over the vocabulary that allows comparing the scores by $\mathbf{v}_i^{\ell}$ w.r.t different tokens.
    
    \item The term $m_i^{\ell}$ is the \textit{dynamic coefficient} of $\mathbf{v}_i^{\ell}$, which is fixed for all tokens for a given input. 
    Thus, these coefficients allow comparing the contribution of value vectors in a specific update.
    
\end{itemize}

\noindent Overall, the scaling factor $\mathbf{e}_w \cdot m_i^{\ell}\mathbf{v}_i^{\ell}$ can be viewed as the effective score given by a value vector $\mathbf{v}_i^{\ell}$ to a token $w$ for a given input.

In the next sections, we use these observations to answer two research questions of (a) What information is encoded in sub-updates and what tokens do they promote? (\S\ref{sec:values_concepts}) and (b) How do FFN updates build the output probability distribution? (\S\ref{sec:values_promotion_fixation})

\begin{figure*}[th!]
  \centering
  \hspace*{\fill}%
  \subfloat{\includegraphics[scale=0.4, trim={0cm 0.2cm 0cm 0.2cm}, clip]{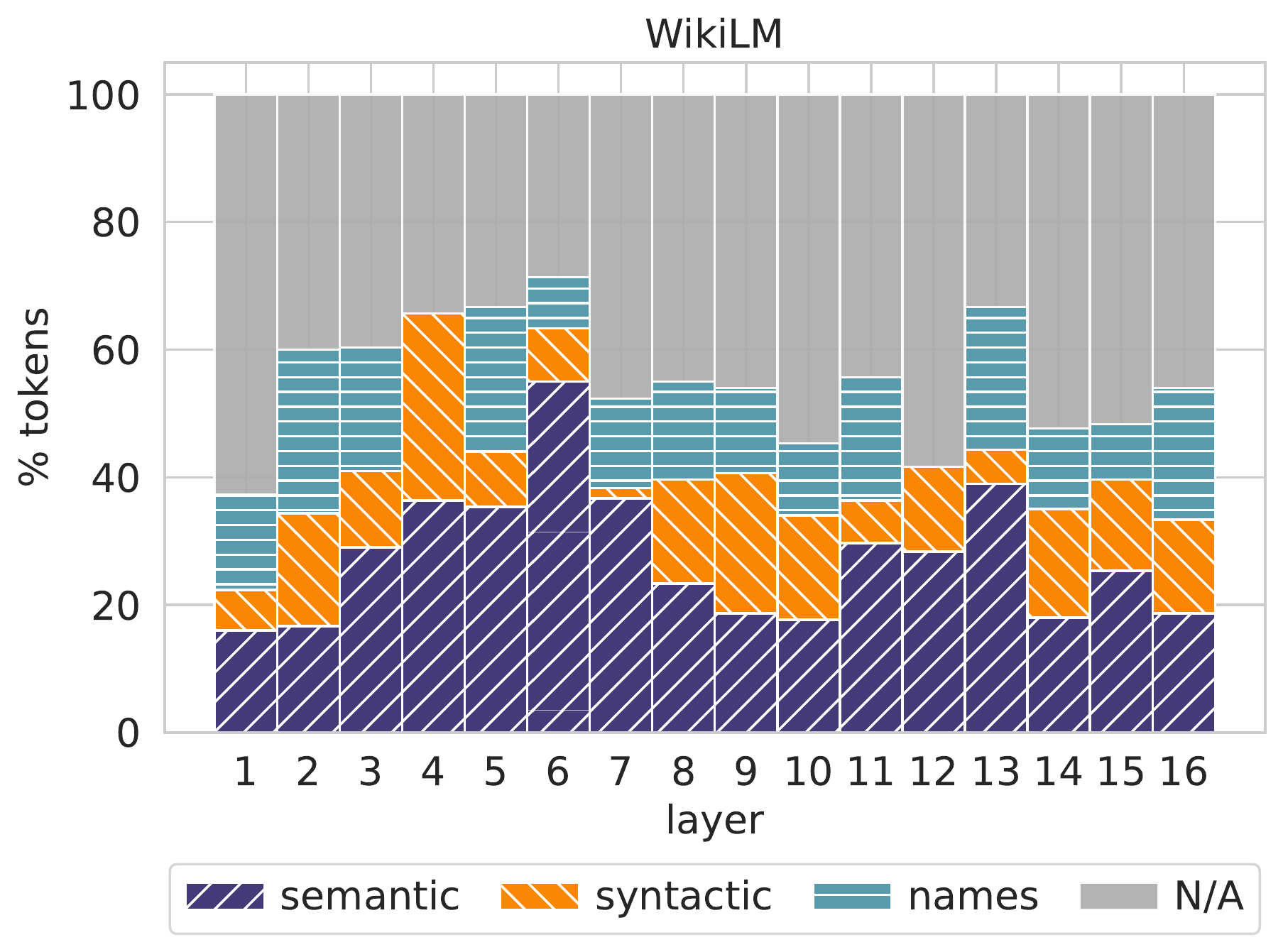}
  }
  \hfill
  \subfloat{\includegraphics[scale=0.4, trim={0cm 0.2cm 0cm 0.2cm}, clip]{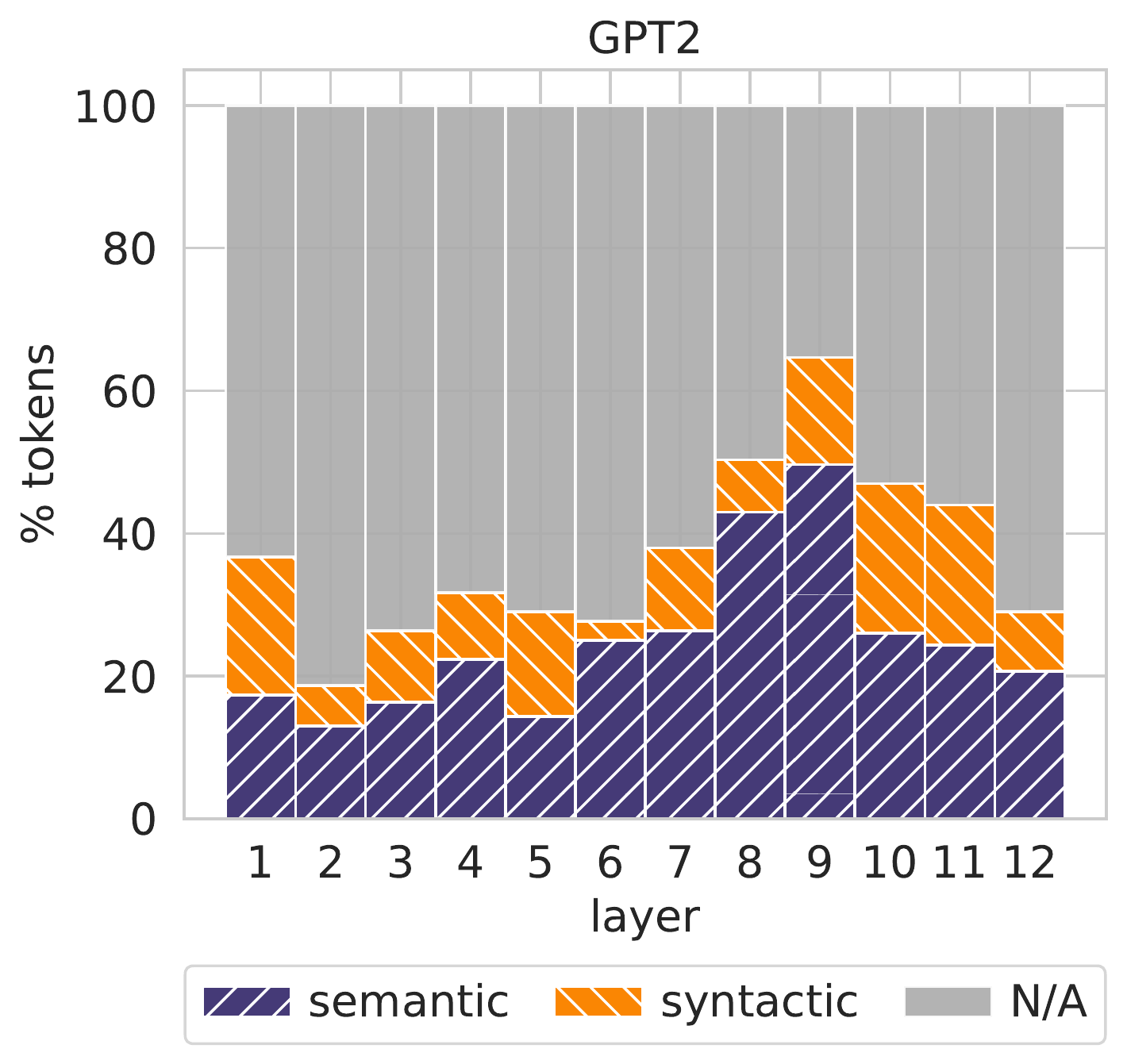}
  }
  \hspace*{\fill}%
  \caption{Portion of top-scoring tokens by value vectors in \wikilm{} and \gpt{}, that were associated with a semantic or syntactic concept, a name, or could not be matched to any concept (``N/A'').}
  \label{figure:concept_frequency}
\end{figure*}

\begin{table*}[ht]
    \setlength\tabcolsep{3.0pt}
    \centering
    \footnotesize
    \begin{tabular}{@{}llp{3.5cm}p{9.8cm}@{}}
        & & \textbf{Concept} & \textbf{Sub-update top-scoring tokens} \\ \toprule
        \multirow{3}{*}{\gpt{}} & \vv$^{3}_{1018}$ & Measurement \tcbox{\footnotesize{semantic}} & \texttt{kg, percent, spread, total, yards, pounds, hours} \\ 
        & \vv$^{8}_{1900}$ &  WH-relativizers \tcbox{\footnotesize{syntactic}} & \texttt{which, whose, Which, whom, where, who, wherein} \\ 
        & \vv$^{11}_{2601}$ & Food and drinks \tcbox{\footnotesize{semantic}} & \texttt{drinks, coffee, tea, soda, burgers, bar, sushi} \\ 
        \midrule
        \multirow{3}{*}{\wikilm{}} & \vv$^{1}_{1}$ & Pronouns \tcbox{\footnotesize{syntactic}} & \texttt{Her, She, Their, her, she, They, their, they, His} \\ 
        & \vv$^{6}_{3025}$ & Adverbs \tcbox{\footnotesize{syntactic}} & \texttt{largely, rapidly, effectively, previously, normally} \\ 
        & \vv$^{13}_{3516}$ & Groups of people \tcbox{\footnotesize{semantic}} & \texttt{policymakers, geneticists, ancestries, Ohioans} \\ 
        \bottomrule
    \end{tabular}
    \caption{Example value vectors in \gpt{} and \wikilm{} promoting human-interpretable concepts.}
    \label{table:example_concepts}
\end{table*}

\section{Sub-Updates Encode Concepts in the Vocabulary Space}
\label{sec:values_concepts}
We evaluate whether projection to the vocabulary provides a meaningful way to ``read'' FFN updates, and the extent to which sub-updates are interpretable based on their projections.
To this end, we manually inspect the top-scoring tokens by value vectors and check if they express interpretable concepts.
Concretely, we consider two representative LMs (details below), and for each vector $\mathbf{v}_i^\ell$ compute a ranking over the vocabulary by sorting the projection $\mathbf{r}_i^\ell$ (\S\ref{sec:hypothesis}). Then, we try to detect patterns in the top-scoring tokens of each value vector.

\paragraph{Concepts Annotation Task.} 
We let experts (NLP graduate students) annotate concepts by identifying common patterns among the top-30 scoring tokens of each value vector. For a set of tokens, the annotation protocol includes three steps of: (a) Identifying patterns that occur in at least 4 tokens, (b) describing each recognized pattern, and (c) classifying each pattern as either \textit{``semantic''} (e.g., mammals), \textit{``syntactic''} (e.g., past-tense verbs), or \textit{``names''}. The last class was added only for \wikilm{} (see below), following the observation that a large portion of the model's vocabulary consists of names. 
Further details, including the complete instructions and a fully annotated example can be found in App.~\ref{sec:appendix_concepts_annotation}.

\paragraph{Models.}
We conduct our experiments over two auto-regressive decoder LMs: The model of~\citet{baevski2018adaptive} (dubbed \wikilm{}), a 16-layer LM trained on the \wikitext{} corpus~\cite{merity2017pointer} with word-level tokenization ($|\mathcal{V}|=267,744$), and \gpt{}~\cite{radford2019language}, a 12-layer
LM trained on \textsc{WebText}~\cite{radford2019language} with sub-word tokenization ($|\mathcal{V}|=50,257$). 
\gpt{} uses the GeLU activation function \cite{hendrycks2016gelu}, while \wikilm{} uses ReLU, and in contrast to \gpt{}, \wikilm{} does not apply layer normalization after FFN updates.
\wikilm{} defines $d=1024, d_m =4096$ and \gpt{} defines $d=768, d_m = 3072$ , resulting in a total of $65k$ and $36k$ value vectors, respectively. For our experiments, we sample 10 random vectors per layer from each model, yielding a total of 160 and 120 vectors to analyze from \wikilm{} and \gpt{}, respectively.

\subsection{Projection of Sub-Updates is Meaningful}
\label{sec:sub_updates_projection}

\paragraph{Real vs. Random Sub-Updates.}
We validate our approach by comparing concepts in top-tokens of value vectors and 10 random vectors from a normal distribution with the empirical mean and standard deviation of the real vectors. 
We observe that a substantially higher portion of top-tokens were associated to a concept in value vectors compared to the random ones (Tab.~\ref{table:projections}): $55.1\%$ vs. $22.7\%$ in \wikilm{}, and $37\%$ vs. $16\%$ in \gpt{}. Also, in both models, the average number of concepts per vector was $>1$ in the value vectors compared to $\sim0.5$ in the random ones.
Notably, no semantic nor syntactic concepts were identified in \wikilm{}'s random vectors, and in \gpt{}, only $4\%$ of the tokens were marked as semantic concepts in the random vectors versus $24.9\%$ in the value vectors.

\paragraph{Updates vs. Sub-Updates.}
We justify the FFN output decomposition by analyzing concepts in the top-tokens of 10 random FFN outputs per layer (Tab.~\ref{table:projections}).
In \wikilm{} (\gpt{}), $39.4\%$ ($46\%$) of the tokens were associated with concepts, but for $19.7\%$ ($34.2\%$) the concept was \textit{``stopwords/punctuation''}. Also, we observe very few concepts ($<4\%$) in the last two layers of \wikilm{}. 
We account this to extreme sub-updates that dominate the layer's output (\S\ref{sec:layer_scores}). Excluding these concepts results in a considerably lower token coverage in projections of updates compared to those of sub-updates: $19.7\%$ vs. $55.1\%$ in \wikilm{}, and $11.8\%$ vs. $36.7\%$ in \gpt{}.

Overall, this shows that projecting sub-updates to the vocabulary provides a meaningful interface to the information they encode. Moreover, decomposing the FFN outputs is necessary for fine-grained interpretation of sub-updates.

\begin{table}[t]
\centering
\footnotesize
        \begin{tabular}[b]{@{}llr@{}}
             & \gpt{} & \wikilm{} \\ 
            \midrule
            FFN sub-updates & \textbf{36.7\%} & \textbf{55.1\%} \\
            ~~~ \textit{$+$ stopwords concepts} & \textit{37\%} & \textit{55.1\%} \\
            \midrule
            Random sub-updates  & 16\%  & 22.7\% \\ 
            FFN updates & 11.8\%  & 19.7\% \\
            ~~~ \textit{$+$ stopwords concepts} & \textit{46\%} & \textit{39.4\%}
            \\ \bottomrule
        \end{tabular}
\caption{Portion of top-scoring tokens associated with a concept, for FFN updates and sub-updates in \wikilm{} and \gpt{}, and for random vectors. For FFN updates/sub-updates, we show results with and without counting concepts marked as stopwords.} 
\label{table:projections}
\end{table}

\subsection{Sub-Update Projections are Interpretable}
\label{sec:sub_updates_concepts}

Fig.~\ref{figure:concept_frequency} shows a breakdown of the annotations across layers, for \wikilm{} and \gpt{}. 
In both models and across all layers, a substantial portion (40\%-70\% in \wikilm{} and 20\%-65\% in \gpt{}) of the top-tokens were associated with well-defined concepts, most of which were classified as \textit{``semantic''}.
Also, we observe that the top-tokens of a single value vector were associated with $1.5$ (\wikilm{}) and $1.1$ (\gpt{}) concepts on average, showing that \textit{sub-updates across all layers encode a small-set of well-defined concepts}. Examples are in Tab.~\ref{table:example_concepts}. 

These findings expand on previous results by \citet{geva-etal-2021-transformer}, who observed that value vectors \textit{in the upper layers} represent next-token distributions that follow specific patterns. Our results, which hold across \textit{all the layers}, suggest that these vectors represent general concepts rather than prioritizing specific tokens.

\paragraph{Underestimation of Concept Frequency.}
In practice, we find that this task is hard for humans,\footnote{A sub-update annotation took $8.5$ minutes on average.} as it requires reasoning over a set of tokens without any context, while tokens often correspond to uncommon words, homonyms, or sub-words. Moreover, some patterns necessitate world knowledge (e.g. \textit{``villages in Europe near rivers''}) or linguistic background (e.g. negative polarity items).
This often leads to undetectable patterns, suggesting that the overall results are an underestimation of the true concept frequency.
Providing additional context and token-related information are possible future directions for improving the annotation protocol.

\paragraph{Implication for Controlled Generation.}
If sub-updates indeed encode concepts, then we can not only interpret their contribution to the prediction, but also \textit{intervene} in this process, by increasing the weights of value vectors that promote tendencies of our choice. We demonstrate this in \S\ref{sec:toxic_language_suppression}.

\section{FFN Updates Promote Tokens in the Output Distribution}
\label{sec:values_promotion_fixation}

We showed that sub-updates often encode interpretable concepts (\S\ref{sec:values_concepts}), but how do these concepts construct the output distribution?
In this section, we show that sub-updates systematically configure the prediction via promotion of candidate tokens.

\subsection{Promoted Versus Eliminated Candidates}
\label{sec:promotion}
Every sub-update $m_i^{\ell} \mathbf{v}_i^{\ell}$ either increases, decreases, or does not change the probability of a token $w$, according to the score $\mathbf{e}_w \cdot m_i^{\ell}\mathbf{v}_i^{\ell}$ (\S\ref{sec:hypothesis}). This suggests three mechanisms by which tokens are pushed to the top of the output distribution -- \textit{promotion}, where sub-updates increase the probability of favorable tokens, \textit{elimination}, where sub-updates decrease candidate probabilities, or a \textit{mixture} of both.
To test what mechanism holds in practice, we analyze the scores sub-updates assign to top-candidate tokens by the representation.
To simplify the analysis, we focus on changes induced by the 10 most dominant sub-updates in each layer, that is, the 10 sub-updates $m_i^{\ell} \mathbf{v}_i^{\ell}$ with the largest contribution to the representation, as measured by $|m_i^{\ell}| \cdot  ||\mathbf{v}_i^{\ell}||$ (see details in App.~\ref{sec:dominant_value_contribution}).

For the experiments, we use a random sample of 2000 examples from the validation set of \wikitext{},\footnote{Data is segmented into sentences \cite{geva-etal-2021-transformer}.} which both \wikilm{} and \gpt{} did not observe during training. As the experiments 
do not involve human annotations, we use a larger \gpt{} model with $L=24, d = 1024, d_m = 4096$.

\begin{table}[t!]
    \setlength\tabcolsep{2.0pt}
    \centering
    \footnotesize
    \begin{tabular}{p{7.5cm}}
    \toprule
    $\mathbf{p}^{\ell}$: \texttt{cow, cat, \textbf{dog}, goat, horse, bear} \\
    $\myhat{\mathbf{p}}^{\ell}$:  \texttt{\textbf{dog}, cat, goat, horse, cow, bear} \\
    \textit{Saturation}: \texttt{dog} is promoted from rank 3 in $\mathbf{p}^{\ell}$ to rank 1 in $\myhat{\mathbf{p}}^{\ell}$, to be the top-candidate until the last layer. \\
    \midrule
    $\mathbf{p}^{\ell}$: \texttt{\textbf{cow}, cat, dog, goat, horse, bear} \\
    $\myhat{\mathbf{p}}^{\ell}$: \texttt{dog, cat, goat, horse, \textbf{cow}, bear} \\ 
    \textit{Elimination}: \texttt{cow} is eliminated from rank 1 in $\mathbf{p}^{\ell}$ to 5 in $\myhat{\mathbf{p}}^{\ell}$. \\
     \bottomrule
    \end{tabular}
    \caption{Example saturation and elimination events, after a FFN update (reference tokens are in bold text).}
    \label{table:event_examples}
\end{table}

\begin{table}[t!]
    \setlength{\tabcolsep}{3.5pt} 
    \centering
    \footnotesize
    \begin{tabular}{@{}llrrr@{}}
    \textbf{Sub-updates} & \textbf{Event} & \textbf{Max.} &  \textbf{Mean} & \textbf{Min.} \\ 
    \midrule
    \multirow{2}{*}{\wikilm{}, dominant} & \textit{saturation} & $1.2$ & $<0.01$ & $-0.8$ \\
    & \textit{elimination} & $0.5$ & $-0.01$ & $-0.5$ \\ \midrule
    \multirow{2}{*}{\wikilm{}, random} & \textit{saturation} & $0.02$ & $<0.01$ & $-0.02$ \\
    & \textit{elimination} & $0.02$ & $<0.01$ & $-0.02$ \\
    \midrule
    \multirow{2}{*}{\gpt{}, dominant} & \textit{saturation} & $8.5$ & $1.3$ & $-4.9$ \\
    & \textit{elimination} & $4.0$ & $0.1$ & $-3.6$ \\ \midrule
    \multirow{2}{*}{\gpt{}, random} & \textit{saturation} & $0.2$ & $0.01$ & $-0.2$ \\
    & \textit{elimination} & $0.1$ & $<0.01$ & $-0.1$ \\
    \bottomrule
    \end{tabular}
    \caption{Maximum, mean, and minimum scores of reference tokens in saturation and elimination events, by the 10 most dominant and 10 random sub-updates.}
    \label{table:prediction_events}
\end{table}

\begin{figure*}[t!]
  \centering
  \hspace*{\fill}%
  \subfloat{\includegraphics[scale=0.37, trim={0cm 0.2cm 0cm 0.2cm}, clip]{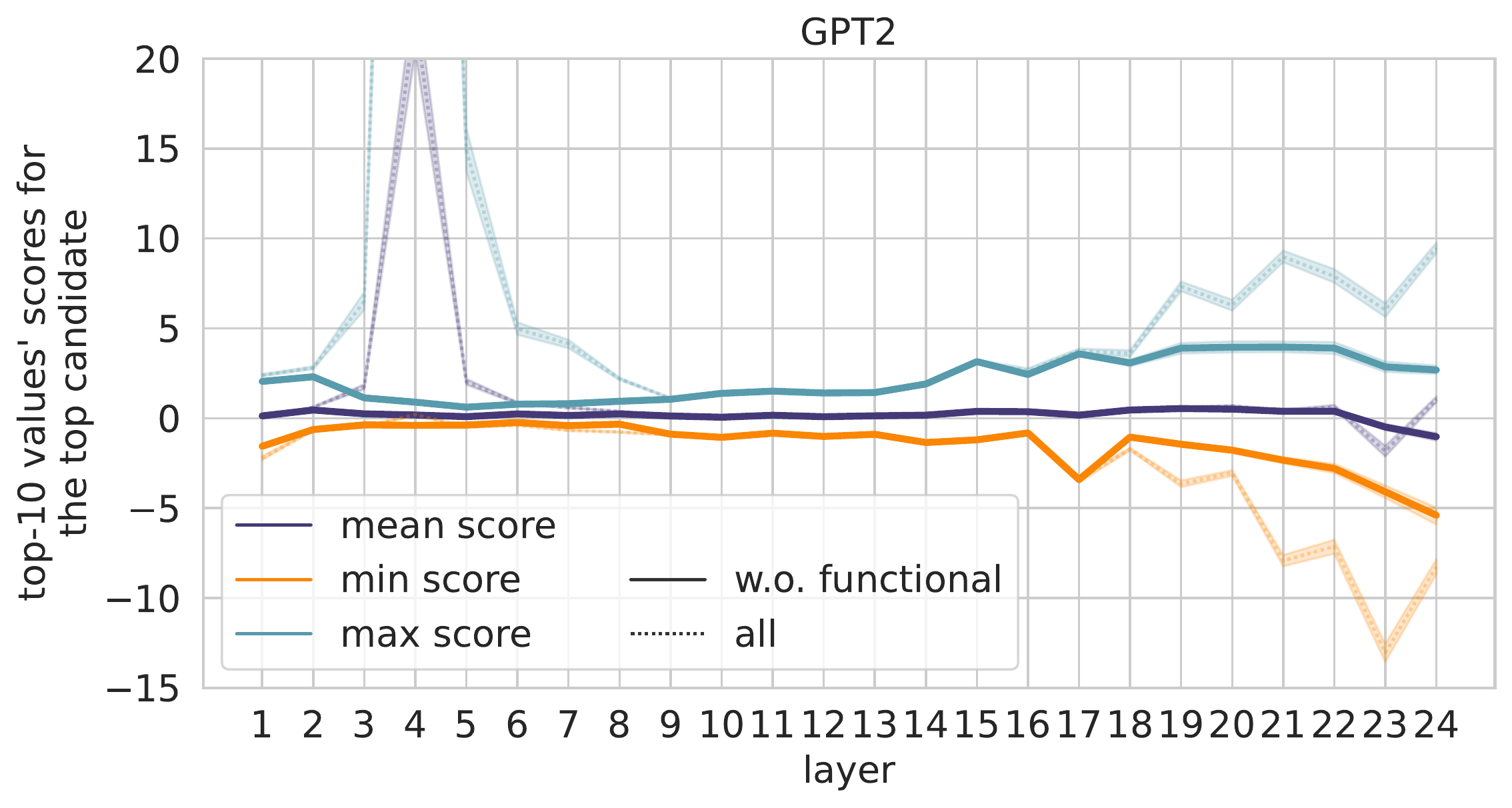}
  }
  \hfill
  \subfloat{\includegraphics[scale=0.37, trim={0cm 0.2cm 0cm 0.2cm}, clip]{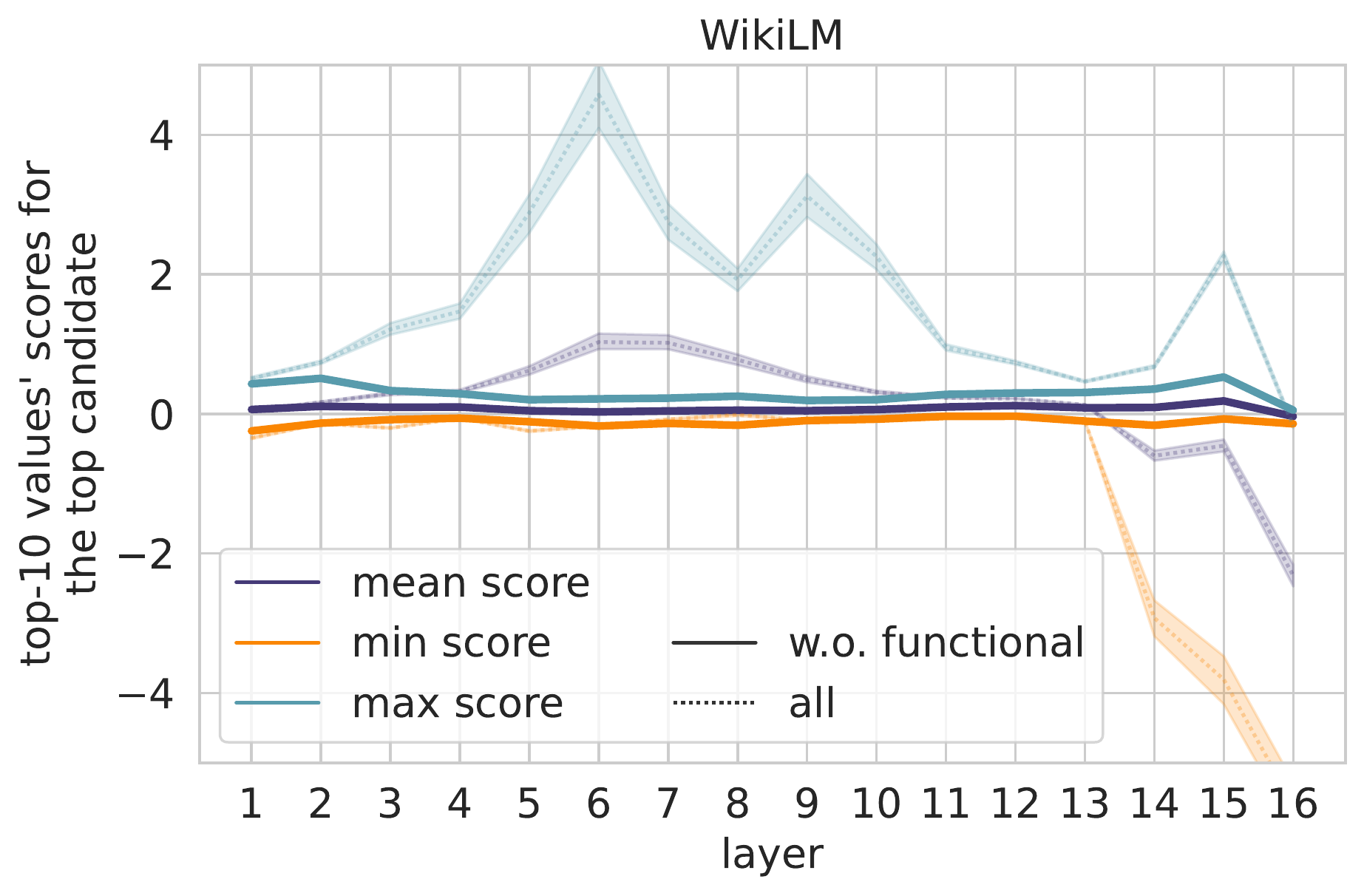}
  }
  \hspace*{\fill}%
  \caption{Mean, maximum and minimum scores assigned by the 10 most dominant sub-updates in each layer to the top-candidate token, in \gpt{} (left) and \wikilm{} (right). Solid (dashed) lines exclude (include) functional value vector groups. The y-axis in both plots is cut for readability, as the max. (min.) scores reach 100 (-6).}
  \label{figure:promotion_fixation}
\end{figure*}

We start by comparing the sub-updates' scores to a reference token in two types of events:
\begin{itemize}
[leftmargin=*,topsep=3pt,itemsep=3pt,parsep=0pt]
    \item \textit{Saturation} (Tab.~\ref{table:event_examples}, up): The update $\mathbf{p}^{\ell} \rightarrow \myhat{\mathbf{p}}^{\ell}$ where the final token predicted by the model (i.e., $w = \text{argmax}(\mathbf{y})$) was promoted to be the top candidate until the last layer. We analyze saturation events induced by the FFN before the last layer, covering 1184 and 1579 events in \wikilm{} and \gpt{}, respectively.
    \item \textit{Elimination} (Tab.~\ref{table:event_examples}, bottom): The update $\mathbf{p}^{\ell} \rightarrow \myhat{\mathbf{p}}^{\ell}$ with the largest increase 
    in the top candidate's rank, i.e. where the top candidate was dropped behind other candidates to have a rank $>1$. Overall, our analysis covers 1909 (\wikilm{}) and 1996 (\gpt{}) elimination events.
\end{itemize}
We compute the mean, maximum, and minimum scores of the reference token by the 10 most dominant sub-updates in each event, and average over all the events.
As a baseline, we compute the scores by 10 random sub-updates from the same layer.

Tab.~\ref{table:prediction_events} shows the results. 
In both models, tokens promoted to the top of the distribution receive
higher maximum scores than tokens eliminated from the top position ($1.2 \rightarrow 0.5$ in \wikilm{} and $8.5 \rightarrow 4.0$ in \gpt{}), indicating they are pushed strongly by a few dominant sub-updates. 
Moreover, tokens eliminated from the top of the distribution receive near-zero mean scores, by both dominant and random sub-updates, suggesting they are not being eliminated directly. 
In contrast to promoted tokens, where the maximum scores are substantially higher than the minimal scores ($1.2$ vs. $-0.8$ in \wikilm{} and $8.5$ vs. $-4.9$ in \gpt{}), for eliminated tokens, the scores are similar in their magnitude ($\pm 0.5$ in \wikilm{} and $4.0$ vs. $-3.6$ in \gpt{}). 
Last, scores by random sub-updates are dramatically lower in magnitude, showing that our choice of sub-updates is meaningful and that higher coefficients translate to greater influence on the output distribution.

\textit{This suggests that FFN updates work in a promotion mechanism, where top-candidate tokens are those being pushed by dominant sub-updates}.

\subsection{Sub-Updates Across Layers}
\label{sec:layer_scores}

To analyze the FFN operation in different layers, we break down the top-candidate scores per layer.
Formally, let $w^{\ell} = \text{argmax}(\mathbf{p}^{\ell})$ be the top candidate at layer $\ell$ (before the FFN update) for a given input, we extract the scores $\mathbf{e}_{w^{\ell}}\cdot m_i^{\ell} \mathbf{v}_i^{\ell}$ by the 10 most dominant sub-updates and compute the mean, minimum and maximum scores over that set.

Fig.~\ref{figure:promotion_fixation}
shows that, in both models, until the last few layers (23-24 in \gpt{} and 14-16 in \wikilm{}), maximum and minimum scores are distributed around non-negative mean scores, with prominent peaks in maximum scores (layers 3-5 in \gpt{} and layers 4-11 in \wikilm{}). This suggests that the token promotion mechanism generally holds across layers.
However, scores diverge in the last layers of both models, with strong negative minimum scores, indicating that the probability of the top-candidate is pushed down by dominant sub-updates.
We next show that these large deviations in positive and negative scores (Fig.~\ref{figure:promotion_fixation}, dashed lines) result from the operation of small sets of functional value vectors.

\paragraph{Extreme Sub-Updates.}
To analyze the extreme FFN updates, we first cluster the value vectors to discover high-level trends.
We use agglomerative clustering~\cite{mullner2011modern} to learn 10k clusters for each model, based on the cosine distance matrix $D$, where $D_{({\ell_1}, {i_1}), ({\ell_2},{i_2})} = 1-cos(\mathbf{v}_{i_1}^{\ell_1},\mathbf{v}_{i_2}^{\ell_2})$,  $
\forall i_1,i_2\in \{1,\cdots,d_m\} ,\; \forall \ell_1,\ell_2 \in \{1,\cdots,L\}
$.\footnote{We experimented with $k=3e^2,1e^3,3e^3,1e^4,3e^4$, and choose $k=1e^4$ based on manual inspection.}
Then, we search for clusters that are frequently active in extreme updates, by 
(a) extracting sub-updates where the scores for the top-candidate pass a certain threshold ($\pm10$ for \gpt{} and $\pm5$ for \wikilm{}), and (b) counting the appearances of each cluster in the layer sub-updates.

In both models, a small set of homogeneous clusters account for the extreme sub-updates shown in Fig.~\ref{figure:promotion_fixation}, which can be divided into two main groups of value vectors: Vectors in the upper layers that promote \textit{generally unlikely} tokens (e.g. rare tokens), and vectors that are spread over all the layers and promote \textit{common} tokens (e.g. stopwords). 
These clusters, which cover only a small fraction of the value vectors (1.7\% in \gpt{} and 1.1\% in \wikilm{}),
are mostly active for examples where the input sequence has $\leq3$ tokens
or when the target token can be easily inferred from the context (e.g. end-of-sentence period), suggesting that these value vectors might configure ``easy'' model predictions. 
More interestingly, the value vectors that promote unlikely tokens can be viewed as \textit{``saturation vectors''}, which propagate the distribution without changing the top tokens. Indeed, these vectors are in the last layers, where often the model already stores its final prediction \cite{geva-etal-2021-transformer}.

\begin{table*}[th!]
    \setlength\tabcolsep{4.0pt}
    \centering
    \footnotesize
    \begin{tabular}{@{}lccccccr@{}}
    \toprule
    \textbf{Model} & \textbf{Toxicity} &  \textbf{Severe} & \textbf{Sexually} & \textbf{Threat} & \textbf{Profanity} & \textbf{Identity} &  \textbf{PPL} \\
     & & \textbf{toxicity} & \textbf{explicit} & & & \textbf{attack} &
    \\ \toprule
    \gpt{} & 58.5\% & 49.2\% & 34.1\% & 16.4\% & 52.5\% & 16.8\% & 21.7 \\ \midrule
    $\uparrow$ 10 Manual Pick & 
    \tcbox{\footnotesize{$\downarrow$47\%}} 30.8\% & 
    \tcbox{\footnotesize{$\downarrow$50\%}} 24.8\% & 
    \tcbox{\footnotesize{$\downarrow$40\% }} 20.4\% & 
    \tcbox{\footnotesize{$\downarrow$63\% }} 6.0\% & 
    \tcbox{\footnotesize{$\downarrow$47\% }} 27.9\% &
    \tcbox{\footnotesize{$\downarrow$48\% }} 8.8\% & 25.3 \\
    
     $\uparrow$ 10 API Graded &
    \tcbox{\footnotesize{$\downarrow$10\% }} 52.7\% & 
    \tcbox{\footnotesize{$\downarrow$11\% }} 44\% & 
    \tcbox{\footnotesize{$\downarrow$3\% }} 33.2\% & 
    \tcbox{\footnotesize{$\downarrow$19\% }} 13.3\% & 
    \tcbox{\footnotesize{$\downarrow$9\% }} 47.6\% & 
    \tcbox{\footnotesize{$\downarrow$9\% }} 15.3\% & 23.8 \\ \midrule
    
    \textsc{SD} & 
    \tcbox{\footnotesize{$\downarrow$37\% }} 37.2\% & 
    \tcbox{\footnotesize{$\downarrow$46\% }} 26.4\% & 
    \tcbox{\footnotesize{$\downarrow$36\% }} 21.7\% & 
    \tcbox{\footnotesize{$\downarrow$52\% }} 7.8\% & 
    \tcbox{\footnotesize{$\downarrow$39\% }} 32\% & 
    \tcbox{\footnotesize{$\downarrow$50\% }} 8.4\% & 23.9 \\
    
    \wordf{} & 
    \tcbox{\footnotesize{$\downarrow$20\% }} 46.9\% & 
    \tcbox{\footnotesize{$\downarrow$34\% }} 32.4\% & 
    \tcbox{\footnotesize{$\downarrow$36\% }} 21.9\% & 
    \tcbox{\footnotesize{$\downarrow$<1\% }} 16.3\% & 
    \tcbox{\footnotesize{$\downarrow$38\% }} 32.3\% & 
    \tcbox{\footnotesize{$\downarrow$13\% }} 14.7\% & - 
    \\ \bottomrule
    \end{tabular}
    \caption{Evaluation results on the challenging subset of \realtoxic{}, showing the percentage of toxic completions for 6 toxicity attributes, as well as language model perplexity (``PPL'').}
    \label{table:toxicity_results}
\end{table*}

\section{Applications}
We leverage our findings for controlled text generation (\S\ref{sec:toxic_language_suppression}) and computation efficiency (\S\ref{sec:early_exit}).

\subsection{Zero-Shot Toxic Language Suppression}
\label{sec:toxic_language_suppression}

LMs are known to generate toxic, harmful language that damages their usefulness \cite{bender2021dangers, mcguffie2020radicalization, wallace-etal-2019-universal}. 
We utilize our findings to create a simple, intuitive method for toxic language suppression.

\paragraph{Method.} If LMs indeed operate in a promotion mechanism, we reason that we can decrease toxicity by ``turning on'' non-toxic sub-updates. 
We find value vectors that promote safe, harmless concepts by extracting the top-tokens in the projections of all the value vectors and either (a) manually searching for vectors that express a coherent set of positive words (e.g. \textit{``safe''} and \textit{``thank''}), or (b) grading the tokens with the Perspective API
and selecting non-toxic value vectors (see details in App.~\ref{sec:toxic_language_suppression_details}). We turn on these value vectors by setting their coefficients to 3, a relatively high value according to Fig.~\ref{figure:promotion_fixation}.
We compare our method with two baselines:
\begin{enumerate}
[leftmargin=*,topsep=3pt,itemsep=3pt,parsep=0pt]
    \item Self-Debiasing (SD) \cite{schick-etal-2021-self}: SD generates a list of undesired words for a given prompt by appending a \textit{self-debiasing input}, which encourages toxic completions, and calculating which tokens are promoted compared to the original prompt. These undesired words' probability are then decreased according to a decay constant $\lambda$, which we set to 50 (default). 
    \item \wordf{}: We prevent \gpt{} from generating words from a list of banned words by setting any logits that would result in a banned word completion to $-\infty$  \cite{gehman-etal-2020-realtoxicityprompts}.
\end{enumerate}

\paragraph{Evaluation.} We evaluate our method on the challenging subset of \realtoxic{} \cite{gehman-etal-2020-realtoxicityprompts}, a collection of 1,225 prompts that tend to yield extremely toxic completions in LMs, using the Perspective API, which grades text according to six toxicity attributes. A score of $>0.5$ indicates a toxic text w.r.t to the attribute. 
Additionally, we compute perplexity to account for changes in LM performance.
We use \gpt{} and,
following \citet{schick-etal-2021-self}, generate continuations of 20 tokens.

\paragraph{Results.} Finding the non-toxic sub-updates manually was intuitive and efficient (taking $<5$ minutes). 
Tab.~\ref{table:toxicity_results} shows that activation of only 10 value vectors (0.01\%) substantially decreases toxicity ($\downarrow$47\%), outperforming both SD ($\downarrow$37\%) and \wordf{} ($\downarrow$20\%). Moreover, inducing sub-updates that promote ``safety'' related concepts is more effective than promoting generally non-toxic sub-updates.
However, our method resulted in a perplexity increase greater than this induced by SD, though the increase was still relatively small.

\subsection{Self-Supervised Early Exit Prediction}
\label{sec:early_exit}
The recent success of transformer-based LMs in NLP tasks has resulted in major production cost increases~\cite{schwartz2020green}, and thus has spurred interest in early-exit methods that reduce the incurred costs~\cite{xu2021survey}.
Such methods often use small neural models to determine when to stop the execution process~\cite{schwartz-etal-2020-right,Elbayad2020Depth-Adaptive,hou2020dynabert,xin-etal-2020-early,xin-etal-2021-berxit,li-etal-2021-cascadebert-accelerating, schuster-etal-2021-consistent}.

In this section, we test our hypothesis that dominant FFN sub-updates can signal a \textit{saturation event} (\S\ref{sec:layer_scores}), to create a simple and effective early exiting method that does not involve any external model training. 
For the experiments, we use \wikilm{}, where saturation events occur across all layers (statistics for \wikilm{} and \gpt{} are in App.~\ref{sec:appendix_early_exit_details}).

\paragraph{Method.} 
We devise a simple prediction rule based on a nearest-neighbours approach, using 10k validation examples from \wikitext{}. 
First, for every example, we map the top-10 dominant sub-updates at each layer to their corresponding clusters. Then, for every layer $\ell$, we split all the sets of clusters at that layer into two sets, $T^{\ell}$ and $N^{\ell}$, based on whether saturation occurred or not (e.g., $T^{5}$ stores all the sets that were active in a saturation event at layer 5).
Given the top-10 clusters of an unseen example at some layer $\ell$, we consider a higher overlap with $T^\ell$ than with $N^{\ell'},\; \forall \ell'>\ell$ as a signal for early exit.
Thus, during inference, we propagate the input example through the layers, and compute at each layer $\ell$ the intersection size between its top-10 active clusters and each of $T^\ell$ and $N^{\ell'},\; \forall \ell'>\ell$. If the average and maximal intersection 
with $T^\ell$ exceeds those with $N^{\ell'},\; \forall \ell'>\ell$, we halt the computation and declare early exiting.\footnote{This is a simplification. We split $N^{\ell}$ by saturation layers and require a bigger intersection with $T^{\ell}$ at all the layers.}

\paragraph{Baselines.}
We train layer-wise binary classifiers over the representation and FFN updates $\mathbf{x}^{\ell}$, $\mathbf{o}^{\ell}$, and $\myhat{\mathbf{x}}^{\ell}$, using logistic regression. As in our method, the labels are determined according to saturation events in the training data (see App.~\ref{sec:appendix_early_exit_details}).
During inference, we execute the computation through the layers, and halt according to the layer classifier.

\paragraph{Evaluation.} Each method is evaluated by \textit{accuracy}, i.e., the portion of examples for which exiting at the predicted layer yields the final model prediction, and by \textit{computation efficiency}, measured by the amount of saved layers for examples with correct prediction. 
We run each method with five random seeds and report the average scores.

\paragraph{Results.} Tab.~\ref{table:early_exit} shows that our method obtains a high accuracy of 94.1\%, while saving 20\% of computation on average without changing the prediction. Moreover, just by observing the dominant FFN sub-updates, it performs on-par with the prediction rules relying on the representation and FFN output vectors. This demonstrates the utility of sub-updates for predicting saturation events, and further supports our hypothesis that FFN updates play a functional role in the prediction (\S\ref{sec:layer_scores}).

\begin{table}[t]
\centering
\footnotesize
\setlength{\tabcolsep}{4pt} 
\setlength{\belowcaptionskip}{-10pt}
        \begin{tabular}[b]{@{}lcr@{}}
            \textbf{Method}            & \textbf{Accuracy} & \textbf{Saved Layers} \\ 
            \midrule
            Binary classifiers using $\mathbf{x}^{\ell}$  & 94.4$\pm$6.4   & \tcbox{18.8\%}3.0$\pm$0.4     \\ 
            Binary classifiers using $\mathbf{o}^{\ell}$  & 92.9$\pm$5.4   & \tcbox{19.4\%}3.1$\pm$0.3   \\
            Binary classifiers using $\myhat{\mathbf{x}}^{\ell}$  & 94.4$\pm$6.4   & \tcbox{18.8\%}3.0$\pm$0.4      \\
            \midrule
            Sub-updates rule & 94.1$\pm$1.4   & \tcbox{20.0\%}3.2$\pm$0.3     
        \end{tabular}
\caption{Early exit evaluation results on \wikilm{}.} 
\label{table:early_exit}
\end{table}

\section{Related Work}
The lack of interpretability of modern LMs has led to a wide interest in understanding their prediction construction process. Previous works mostly focused on analyzing the evolution of hidden representations across layers~\cite{voita-etal-2019-bottom}, and probing the model with target tasks \cite{yang-etal-2020-sub,clark-etal-2019-bert,tenney-etal-2019-bert,saphra-lopez-2019-understanding}. In contrast, our approach aims to interpret the model parameters and their utilization in the prediction process.

More recently, a surge of works have investigated the knowledge captured by the FFN layers \cite{da2021analyzing, jiang-etal-2020-know,dai2021knowledge, yao2022kformer,meng2022locating,singh-etal-2020-bertnesia}. These works show that the FFN layers store various types of knowledge, which can be located in specific neurons and edited. Unlike these works, we focus on the FFN outputs and their contribution in the prediction construction process.

Last, our interpretation of FFN outputs as updates to the output distribution relates to recent works that interpreted groups of LM parameters in the discrete vocabulary space \cite{geva-etal-2021-transformer, khashabi2021prompt}, or viewed the representation as an information stream \cite{elhage2021mathematical}.

\section{Conclusions}

Understanding the inner workings of transformers is valuable for explainability to end-users, for debugging predictions, for eliminating undesirable behavior, and for understanding the strengths and limitations of NLP models. The FFN is an understudied core component of transformer-based LMs, which we focus on in this work.

We study the FFN output as a linear combination of parameter vectors, termed values, and the mechanism by which these vectors update the token representations. We show that value vectors often encode human-interpretable concepts and that these concepts are promoted in the output distribution. 

Our analysis of transformer-based LMs provides a more detailed understanding of their internal prediction process, and suggests new research directions for interpretability, control, and efficiency, at the level of individual vectors.

\section*{Limitations}

Our study focused on the operation of FFN layers in building model predictions. Future work should further analyze the interplay between these layers and other components in the network, such as attention-heads. 

In our analysis, we decomposed the computation of FFN layers into smaller units, corresponding to single value vectors. 
However, it is possible that value vectors are compositional in the sense that combinations of them may produce new meanings. Still, we argue that analyzing individual value vectors is an important first step, since (a) the space of possible combinations is exponential, and (b) our analysis suggests that aggregation of value vectors is less interpretable than individual value vectors (\S\ref{sec:sub_updates_projection}). Thus, this approach opens new directions for interpreting the contribution of FFN layers to the prediction process in transformer LMs.

In addition, we chose to examine the broad family of decoder-based, auto-regressive LMs, which have been shown to be extremely effective for many NLP tasks, including few- and zero-shot tasks \cite{wang2022language}. 
While these models share the same building blocks of all transformer-based LMs, it will be valuable to ensure that our findings still hold for other models, such as encoder-only LMs (e.g. RoBERTa \cite{liu2019roberta}) and models trained with different objective functions (e.g. masked language modeling \cite{devlin2018bert}).


Finally, our annotation effort was made for the evaluation of our hypothesis that sub-updates encode human-interpretable concepts. Scaling our annotation protocol would enable a more refined map of the concepts, knowledge and structure captured by LMs. 
Furthermore, since our concept interpretation approach relies on manual inspection of sets of tokens, its success might depend on the model's tokenization method. In this work, we analyzed models with two different commonly-used tokenizers, and future research could verify our method over other types of tokenizations as well.




\section*{Ethics Statement}
Our work in understanding the role that single-values play in the inference that transformer-based LMs perform potentially improves their transparency, while also providing useful control applications that save energy (early-exit prediction) and increase model harmlessness (toxic language suppression). It should be made clear that our method for toxic language suppression only reduces the probability of toxic language generation and does not eliminate it. As such, this method (as well as our early-exit method) should not be used in the real world without further work and caution. 

More broadly, our work suggests a general approach for modifying LM predictions in particular directions, by changing the weights of FFN sub-updates.
While this is useful for mitigating biases, it also has the potential for abuse. It should be made clear that, as in the toxic language suppression application, our approach does not modify the information encoded in LMs, but only changes the intensity in which this information is exposed in the model's predictions.
Moreover, our work primarily proposes an interpretation for FFN sub-updates, which also could be used to identify abusive interventions.
Regardless, we stress that LMs should not be integrated into critical systems without caution and monitoring.

\section*{Acknowledgements}
We thank Shauli Ravfogel, Tal Schuster, and Jonathan Berant for helpful feedback and constructive suggestions.
This project has received funding from the Computer Science Scholarship granted by the Séphora Berrebi Foundation, the PBC fellowship for outstanding PhD candidates in Data Science, and the European Research Council (ERC) under the European Union's Horizon 2020 research and innovation programme, grant agreement No. 802774 (iEXTRACT).

\bibliography{all,anthology}
\bibliographystyle{acl_natbib}

\clearpage

\appendix

\section{Appendix}
\label{sec:appendix}

\begin{figure}[t]
    \centering
    \includegraphics[scale=0.38]{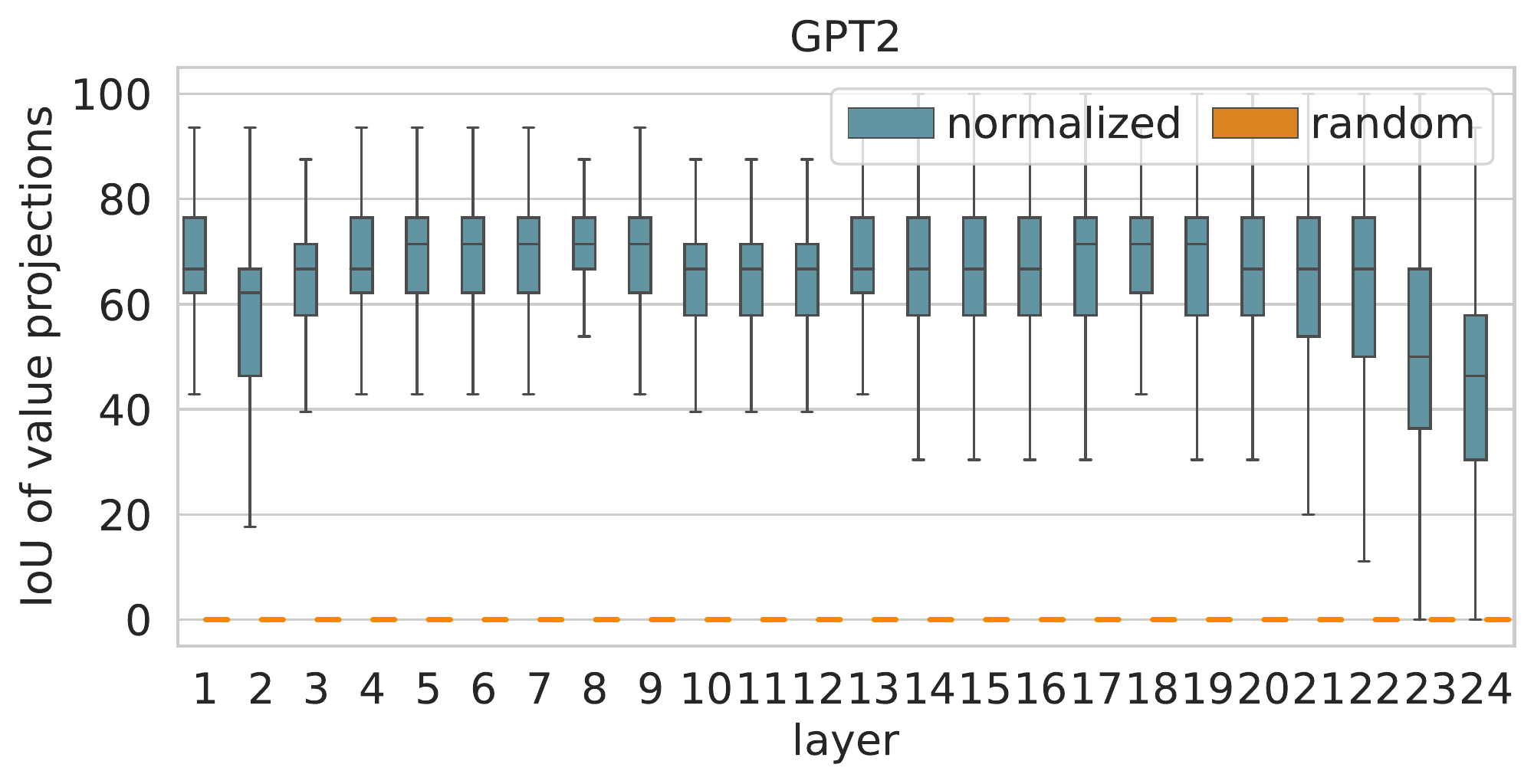}
    \caption{Similarity of projections to $E$, of \gpt{} value vectors with and without layer normalization, and of value vectors and randomly-initialized vectors. We measure similarity of the top-30 tokens in each projection with IoU.}
    \label{figure:gpt2_value_projections_iou}
\end{figure}

\subsection{Value Vectors Projection Method}
\label{sec:value_layer_norm}

Our interpretation method of sub-updates is based on directly projecting value vectors to the embedding matrix, i.e. for a value $\mathbf{v}$ and embedding matrix $E$, we calculate $E \mathbf{v}$ (\S\ref{sec:values_concepts}).
However, in some LMs like \gpt{}, value vectors in each layer are added to the token representation followed by a layer normalization (LN) \cite{ba2016layer}. This raises the question whether ``reading'' vectors that are normalized in the same manner as the representation would yield different concepts.

To test that, we compare the top-30 scoring tokens by $E \mathbf{v}_i^{\ell}$ and by $E \cdot \text{LayerNorm}(\mathbf{v}_i^{\ell})$, for $i=1,...,d_m$ and $\ell=1,...,L$, using Intersection over Union (IoU). As a baseline, we also compare $E \mathbf{v}_i^{\ell}$ with random vectors, initialized from a normal distribution with the empirical mean and standard deviation of the value vectors.
Fig.~\ref{figure:gpt2_value_projections_iou} shows that LN does not change the projection substantially, with an overlap of $64.5\%$ of the top-30 tokens on average, suggesting that the same concepts are promoted in both cases. This is in contrast to random values, which produce a $\sim0\%$ overlap on average.

\begin{figure*}[ht]
    \centering
    \includegraphics[scale=0.75, trim=100 520 100 50]{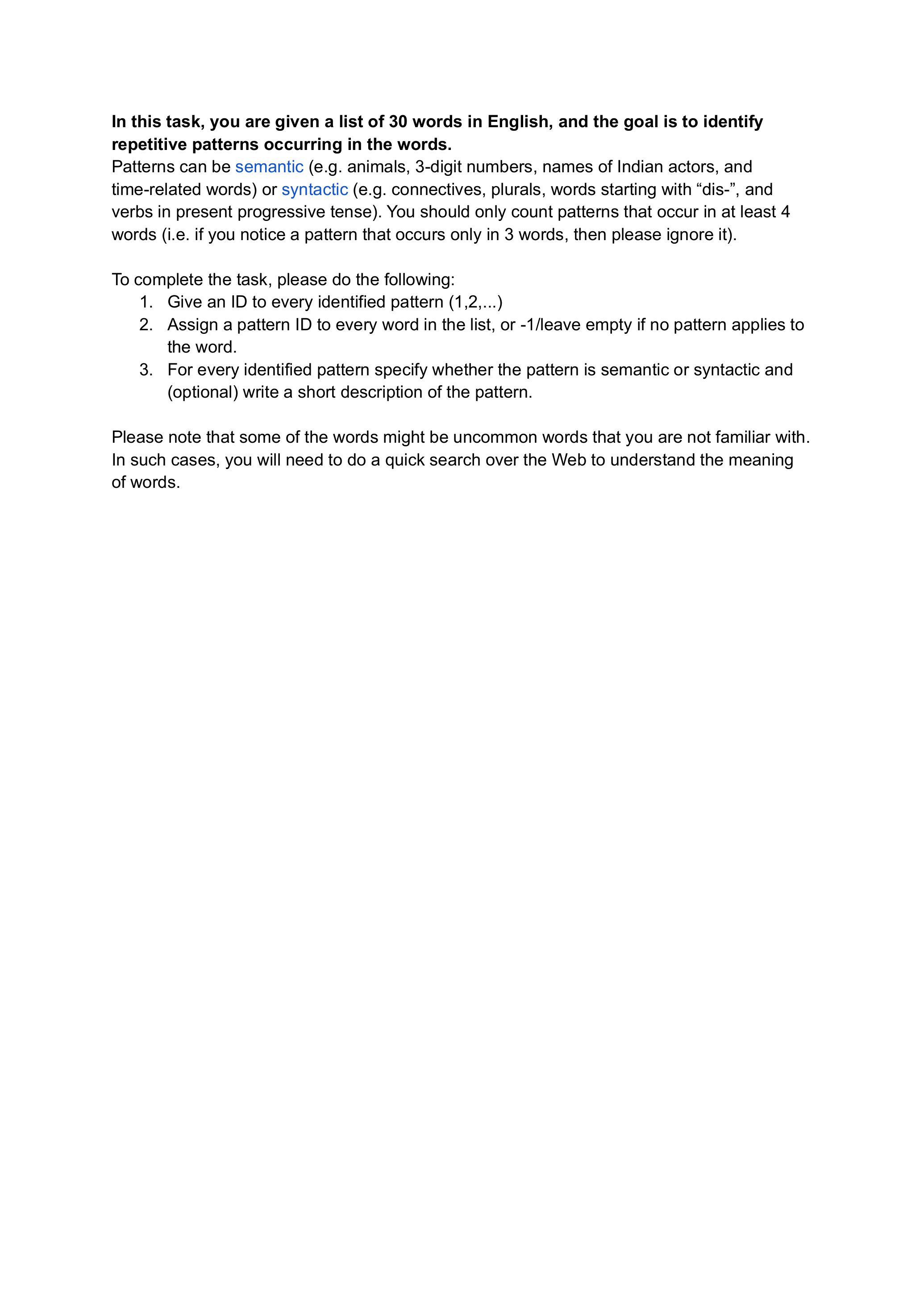}
    \caption{Annotation instructions for the concepts identification task.}
    \label{figure:annotation_instructions}
\end{figure*}

\begin{table*}[th]
    \centering
    \footnotesize
    \begin{tabular}{|c|p{1.9cm}|p{11.5cm}|}
        \hline 
         patterns & word & description \\ \hline
1 & front & the side that is forward or prominent \\
1 & ahead & having the leading position or higher score in a contest \\
1 & forward & the person who plays the position of forward in certain games, such as basketball, soccer, or hockey \\
1 & preceded & be earlier in time; go back further \\
1 & Before & earlier in time; previously \\
1 & before & earlier in time; previously \\
1 & rear & the back of a military formation or procession \\
1 & fore & front part of a vessel or aircraft \\
2 & Name & a language unit by which a person or thing is known \\
1 & Past & the time that has elapsed \\
1 & prior & the head of a religious order; in an abbey the prior is next below the abbot \\
1 & anterior & a tooth situated at the front of the mouth \\
1 & upperparts & standard terms for unambiguous description of relative placement of body parts \\
1 & lead & an advantage held by a competitor in a race \\
1 & backwards & at or to or toward the back or rear \\
1 & aft & (nautical, aeronautical) situated at or toward the stern or tail \\
1 & preceding & be earlier in time; go back further \\
1 & upstream & in the direction against a stream's current \\
 & hind & any of several mostly spotted fishes that resemble groupers \\
1 & posterior & the fleshy part of the human body that you sit on \\
 & Etymology & a history of a word \\
1 & Pre & Wikimedia disambiguation page \\
 & chin & the protruding part of the lower jaw \\
1 & north & the region of the United States lying to the north of the Mason-Dixon line \\
1 & east & the cardinal compass point that is at 90 degrees \\
2 & surname & the name used to identify the members of a family (as distinguished from each member's given name) \\
1 & Then & that time; that moment \\
2 & name & a language unit by which a person or thing is known \\
1 & northbound & moving toward the north \\
1 & leading & thin strip of metal used to separate lines of type in printing \\ \hline 
\multicolumn{3}{}{*} \\ \hline
pattern id & description (optional) & semantic/syntactic \\ \hline
1 & positions/ directions & semantic \\
2 & naming & semantic \\ \hline
    \end{tabular}
    \caption{An example annotation spreadsheet of the top-tokens by the value vector $\mathbf{u}_{1090}^6$ in \wikilm{}.}
    \label{table:annotation_example}
\end{table*}

\subsection{Concepts Annotation}
\label{sec:appendix_concepts_annotation}

We analyze the concepts encoded in sub-updates, by projecting their corresponding value vectors to the embedding matrix and identifying repeating patterns in the top-scoring 30 tokens (\S\ref{sec:hypothesis}).
Pattern identification was performed by experts (NLP graduate students), following the instructions presented in Fig.~\ref{figure:annotation_instructions}. Please note these are the instructions provided for annotations of \wikilm{}, which uses word-level tokenization. Thus, the terms ``words'' and ``tokens'' are equivalent in this case.

For value vectors in \wikilm{}, which uses a word-level vocabulary with many uncommon words, we additionally attached a short description field for each token that provides context about the meaning of the word. For the description of a token $w$, we first try to extract the definition of $w$ from Wordnet.\footnote{We use the NLTK python package.}
If $w$ does not exist in Wordnet, as often happens for names of people and places, we then search for $w$ in Wikipedia\footnote{Using the wptools package \url{https://pypi.org/project/wptools/}.} and extract a short (possibly noisy) description if the query was successful.
A complete annotation example Tab.~\ref{table:annotation_example}.

\subsection{Sub-Update Contribution in FFN Outputs}
\label{sec:dominant_value_contribution}
In this section, we justify our choice along the paper of looking at the top-10 dominant sub-updates. 
The contribution of a sub-update $m_i^{\ell} \mathbf{v}_i^{\ell}$ to the FFN output is:
$$
\text{contrib}(m_i^{\ell}\mathbf{v}_i^{\ell}) := \frac{|m_i^{\ell}| ||\mathbf{v}_i^{\ell}||}{\sum_{j=1}^{d_m}|m_{j}^{\ell}| ||\mathbf{v}_j^{\ell}||},
$$ 
namely, its relative weight compared to the overall sum of weights of all sub-updates. The overall contribution of the top-10 dominant sub-updates is computed by summing their contributions. 
Note that we take the absolute value of the coefficients $|m_i^{\ell}|$, since some activation functions (e.g. GeLU~\cite{hendrycks2016gelu} in \gpt{}), can result in negative values of $m_i^{\ell}$.

Empirically, we observe that in some cases sub-updates with negative coefficients do appear as part of the 10 most dominant sub-updates in \gpt{}.
We further attribute this to the success of GeLU in transformer models~\cite{Shazeer2020GLUVI}, as it increases the expressiveness of the model by allowing reversing the scores value vectors induce over the vocabulary.

\begin{figure*}[th!]
  \centering
  \hspace*{\fill}%
  \subfloat{\includegraphics[scale=0.4]{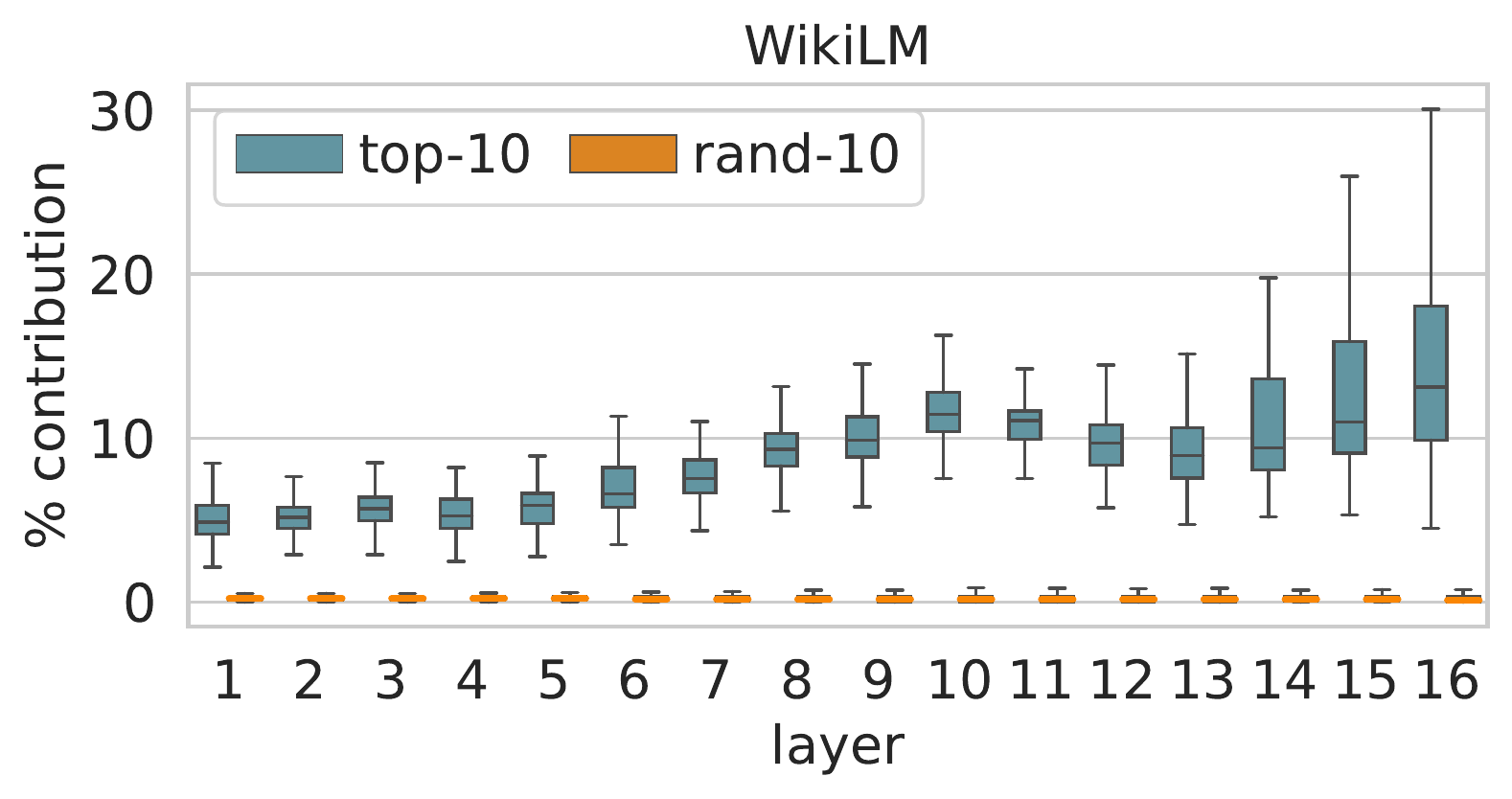}
  }
  \hfill
  \subfloat{\includegraphics[scale=0.4]{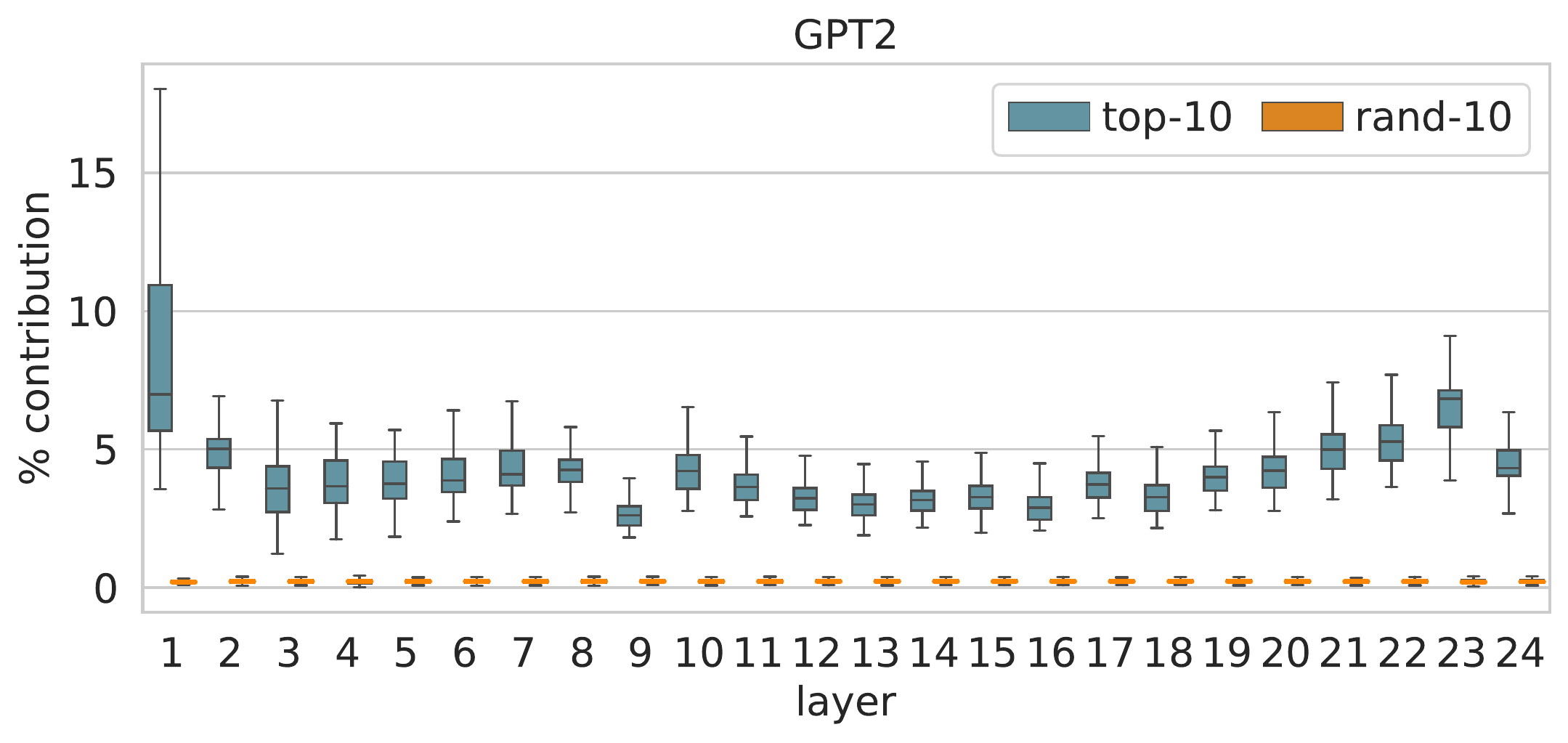}
  }
  \hspace*{\fill}%
    \caption{Relative contribution to the FFN output of the 10 most dominant and 10 random sub-updates in each layer, of \wikilm{} (left) and \gpt{} (right).}
  \label{figure:value_contribution}
\end{figure*}

Fig.~\ref{figure:value_contribution} depicts the contribution of the top-10 dominant sub-updates per layer for \wikilm{} and \gpt{}, using 2000 random examples from the \wikitext{} validation set. Clearly, for all the layers, the contribution of the dominant sub-updates exceeds the contribution of random sub-updates. 
Observe that, even though they cover only 0.24\% of the value vectors, the contribution of dominant sub-updates is typically around 5\%, and in some layers (e.g. layers 8-16 in \wikilm{} and layer 1 in \gpt{}) it reaches over 10\% of the total contribution. 
This demonstrates that analyzing the top-10 dominant sub-updates can shed light on the way predictions are built through the layers.

\subsection{Toxic Language Suppression Details}
\label{sec:toxic_language_suppression_details}
The 10 manually selected value vectors were found by searching for non-toxic words, such as \textit{``safe''} and \textit{``peace''}, among the top-30 tokens in the vector projections to the vocabulary. We selected a small set of 10 value vectors whose top-scoring tokens were coherent and seemed to promote different kinds of non-toxic tokens. 
The list of manually picked vectors is provided in Tab.~\ref{table:toxic_suppression_values}.
Importantly, the search process of all vectors was a one-time effort that took $<5$ minutes in total. We chose the value vectors in a greedy-manner, without additional attempts to optimize our choice.

To select 10 non-toxic value vectors based on an automatic toxicity metric, we used the Perspective API. Concretely, we concatenated the top-30 tokens by each value vector and graded the resulting text with the toxicity score produced by the API. Then, we sampled 10 random vectors with a toxicity score $<0.1$ (a score of $<0.5$ indicates a non-toxic text).

\begin{table*}[ht]
    \centering
    \footnotesize
    \begin{tabular}{lp{9cm}} \textbf{Value} & \textbf{Top-10 Tokens} \\ \toprule
        \multirow{2}{*}{\vv$^{14}_{1853}$} & \texttt{transparency, disclosure, clearer, parency, iquette, humility, modesty, disclosures, accountability, safer} \\ \hline 

        \multirow{2}{*}{\vv$^{15}_{73}$} & \texttt{respectful, honorable, healthy, decent, fair, erning, neutral, peacefully, respected, reconc} \\ \hline 

        \multirow{2}{*}{\vv$^{15}_{1395}$} & \texttt{safe, neither, safer, course, safety, safe, Safe, apologize, Compact, cart} \\ \hline 

        \multirow{2}{*}{\vv$^{16}_{216}$} & \texttt{refere, Messages, promises, Relations, accept, acceptance, Accept, assertions, persistence, warn} \\ \hline 
        \multirow{2}{*}{\vv$^{17}_{462}$} & \texttt{should, should, MUST, ought, wisely, Should, SHOULD, safely, shouldn, urgently} \\ \hline 

        \multirow{2}{*}{\vv$^{17}_{3209}$} & \texttt{peaceful, stable, healthy, calm, trustworthy, impartial, stability, credibility, respected, peace} \\ \hline 
        
        \multirow{2}{*}{\vv$^{17}_{4061}$} & \texttt{Proper, proper, moder, properly, wisely, decency, correct, corrected, restraint, professionalism} \\ \hline 

        \multirow{2}{*}{\vv$^{18}_{2921}$} & \texttt{thank, THANK, thanks, thank, Thank, apologies, Thank, thanks, Thanks, apologise} \\ \hline 
        
        \multirow{2}{*}{\vv$^{19}_{1891}$} & \texttt{thanks, thank, Thanks, thanks, THANK, Thanks, Thank, Thank, thank, congratulations} \\ \hline 
        
        \multirow{2}{*}{\vv$^{23}_{3770}$} & \texttt{free, fit, legal, und, Free, leg, pless, sound, qualified, Free} \\ 
         \bottomrule
    \end{tabular}
    \caption{The 10 manually picked value vectors used for toxic language suppression and the top-10 tokens in their projection to the vocabulary. Repetitions in the projections are a result of special characters not being shown. These vectors were found by manually searching for non-toxic words such as \textit{``safe''} and \textit{``peace''} in the projections to the vocabulary.}
    \label{table:toxic_suppression_values}
\end{table*}

\subsection{Early Exit Details}
\label{sec:appendix_early_exit_details}
This section provides further details and analysis regarding our early exit method and the baselines we implemented.

\paragraph{Method Implementation.}
We consider 90\% of the 10k examples for constructing $T^\ell$ and $N^\ell$, and the remaining 10\% examples are considered as the testing set. We used $k=2e^2$ to cluster the top-10 dominant value vectors, but observed that other $k$ values yielded similar results.

\paragraph{Baselines' Implementation.}
We train each binary classifier using 8k training examples, based on the standardized forms of each feature vector. We considered a hyperparameter sweep, using 8-fold cross-validation, with $l2$ or $l1$ regularization (lasso \cite{tibshirani1996regression} or ridge \cite{hoerl1970ridge}), regularization coefficients ${C\in\{1e^{-3},1e^{-2},1e^{-1},1,1e^{1},1e^{2},1e^{3}\}}$, and took the best performing model for each layer. We also used a inversely proportional loss coefficient according to the class frequencies.

In order to achieve high accuracy, we further calibrate a threshold per classifier for reaching the maximal F$_1$ score for each layer. This calibration is done after training each classifier, over a set of 1000 validation examples. 

\paragraph{Frequency of Saturation Events.}
\begin{table}[htb]
\small
\centering
\begin{tabular}{c c | cc} \toprule
             \textbf{Layer} & \textbf{\% Examples}& \textbf{Layer} & \textbf{\% Examples} \\ \midrule
             
             1	&	6.70	&	9	&	2.96	\\
            2	&	5.25	&	10	&	3.78	\\
            3	&	13.74	&	11	&	4.74	\\
            4	&	3.13	&	12	&	7.45	\\
           5	&	1.02	&	13	&	10.79	\\
           6	&	1.07	&	14	&	9.88	\\
           7	&	1.86	&	15	&	9.81	\\
           8	&	2.60	&	16	&	15.22	\\
              
             \bottomrule
            
\end{tabular}
\caption{The percentage of saturation events per layer using \wikilm{}, for the \wikitext{} validation set.}
\label{table:fixation_events_wikilm}
\end{table}

\begin{table}[htb]
\small
\centering
\begin{tabular}{c c | cc} \toprule
             \textbf{Layer} & \textbf{\% Examples}& \textbf{Layer} & \textbf{\% Examples} \\ \midrule
             
             1	&	2.21	&	13	&	1.24	\\
            2	&	0.77	&	14	&	1.62	\\
            3	&	1.06	&	15	&	2.37	\\
            4	&	0.74	&	16	&	2.72	\\
           5	&	0.85	&	17	&	2.99	\\
           6	&	0.83	&	18	&	3.80	\\
           7	&	0.83	&	19	&	4.15	\\
           8	&	0.72	&	20	&	5.21	\\
           9	&	0.93	&	21	&	5.67	\\
           10	&	0.99	&	22	&	9.31	\\
           11	&	1.16	&	23	&	14.52	\\
           12	&	1.32	&	24	&	34.15	\\
              
             \bottomrule
            
\end{tabular}
\caption{The percentage of saturation events per layer using \gpt{}, for the \wikitext{} validation set.}
\label{table:fixation_events_gpt}
\end{table}

We investigate the potential of performing early exit for \wikilm{} and \gpt{}. 
Tab.~\ref{table:fixation_events_wikilm} and~\ref{table:fixation_events_gpt} depict the frequency of saturation events per layer, considering 10k examples from the \wikitext{} validation set, for \wikilm{} and \gpt{}, respectively. 
In \gpt{}, 34.15\% of the examples require the full computation using all the model layers, while for \wikilm{}, this holds for only 15.22\% of the examples. 
Notably, early fixation events in \gpt{} are less common than in \wikilm{}, possibly due to the larger number of layers the prediction construction is spread over. 
Hence, we use \wikilm{} for our experiments, as it has significantly higher computation saving potential, as well as more saturation events per layer.

\end{document}